\documentclass{article}

\PassOptionsToPackage{numbers, compress}{natbib}

\usepackage[final]{neurips_2024}

\usepackage[utf8]{inputenc} 
\usepackage[T1]{fontenc}    
\usepackage{hyperref}       
\usepackage{url}            
\usepackage{booktabs}       
\usepackage{amsfonts}       
\usepackage{nicefrac}       
\usepackage{microtype}      
\usepackage{hyperref}
\usepackage{url}
\usepackage{booktabs}
\usepackage[table]{xcolor}
\usepackage{enumitem}
\usepackage{array}
\usepackage{caption}
\usepackage{amsmath}
\usepackage{graphicx} 
\usepackage{multicol}

\definecolor{groupbg}{RGB}{255,229,204}
\newcommand{\grp}{\rowcolor{groupbg}}

\title{Xiaomi MiMo-VL-Miloco Technical Report}

\author{%
  MiLM-Plus Xiaomi
}

\begin{document}

\maketitle

\begin{abstract}
We open-source \textbf{MiMo-VL-Miloco-7B} and its quantized variant \textbf{MiMo-VL-Miloco-7B-GGUF}, a pair of home-centric vision-language models that achieve strong performance on both home-scenario understanding and general multimodal reasoning. Built on the MiMo-VL-7B backbone, MiMo-VL-Miloco-7B is specialized for smart-home environments, attaining leading F1 scores on gesture recognition and common home-scenario understanding, while also delivering consistent gains across video benchmarks such as Video-MME, Video-MMMU, and Charades-STA, as well as language understanding benchmarks including MMMU-Pro and MMLU-Pro. In our experiments, MiMo-VL-Miloco-7B outperforms strong closed-source and open-source baselines on home-scenario understanding and several multimodal reasoning benchmarks. To balance specialization and generality, we design a two-stage training pipeline that combines supervised fine-tuning with reinforcement learning based on Group Relative Policy Optimization, leveraging efficient multi-domain data. We further incorporate chain-of-thought supervision and token-budget-aware reasoning, enabling the model to learn knowledge in a data-efficient manner while also performing reasoning efficiently. Our analysis shows that targeted home-scenario training not only enhances activity and gesture understanding, but also improves text-only reasoning with only modest trade-offs on document-centric tasks. Model checkpoints, quantized GGUF weights, and our home-scenario evaluation toolkit are publicly available at \href{https://github.com/XiaoMi/xiaomi-mimo-vl-miloco}{https://github.com/XiaoMi/xiaomi-mimo-vl-miloco} to support research and deployment in real-world smart-home applications.

\begin{figure}[h!]
    \centering
    \includegraphics[width=\textwidth]{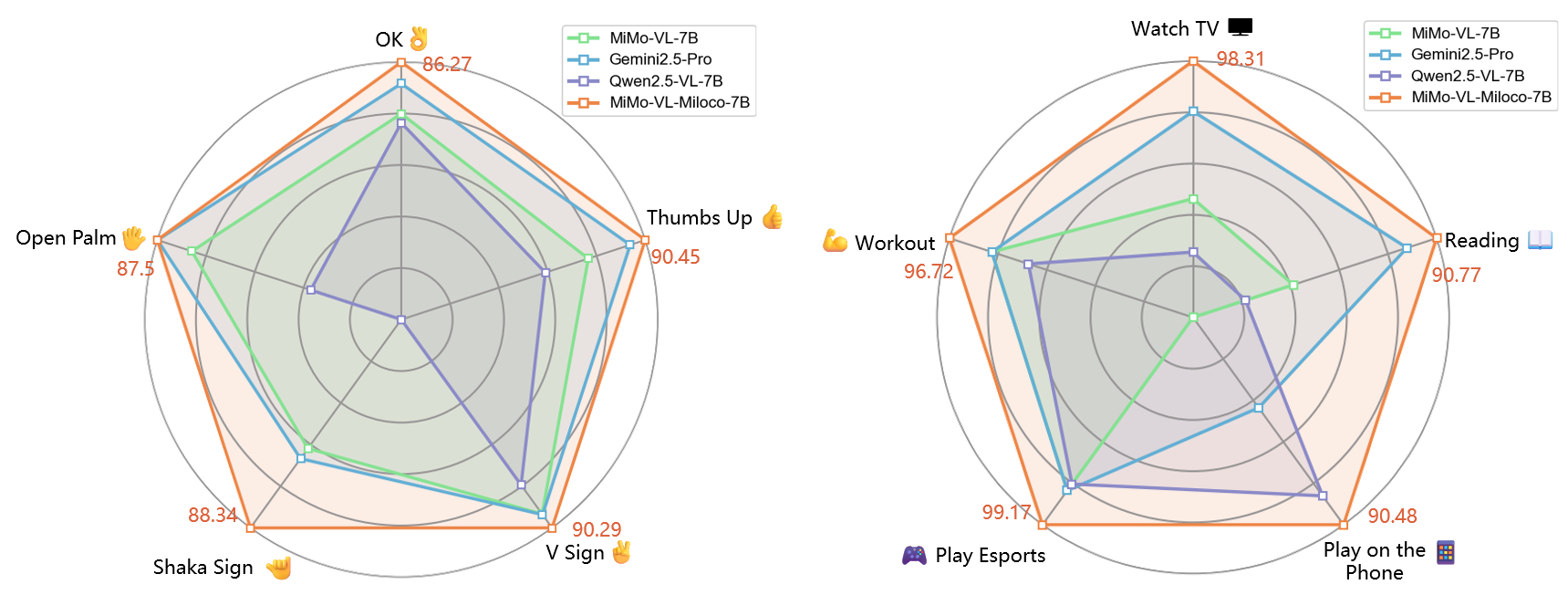}
    \caption{Per-category F1 score comparison on home-scenario gesture recognition (left) and daily-activity recognition (right) benchmarks for MiMo-VL-Miloco-7B and baseline models (MiMo-VL-7B-SFT-2508, Gemini-2.5-Pro, and Qwen2.5-VL-7B). MiMo-VL-Miloco-7B attains the highest F1 scores on most gestures (OK, Thumbs Up, V Sign, Shaka Sign, Open Palm) and activities (Watch TV, Reading, Play on the Phone, Play Esports, Workout), with the annotated values indicating its per-category F1 scores.}
    \label{fig:radar}
\end{figure}

\end{abstract}

\newpage

\section{Introduction}

Recently, vision-language models and, more broadly, multimodal large language models have rapidly become the core infrastructure for modern AI systems, enabling agents to perceive visual scenes, reason over images and videos, and interact with the physical and digital worlds in a unified way \cite{alayrac2022flamingo,liu2023visual,radford2021learning,yin2024survey,li2023blip,li2025revisor, li2025imove}. In the smart-home domain, these capabilities are especially critical: to move beyond simple ``if-this-then-that'' rules, home systems must be able to understand what people are doing, interpret fine-grained visual cues, and make safe, context-aware decisions in real time \cite{bouchabou2021survey, diraco2023human, yu2022deep}. In particular, privacy and latency constraints call for compact models that can run locally on edge devices instead of relying solely on cloud inference \cite{dhar2021survey, surianarayanan2023survey}.

In this work, we present \textbf{MiMo-VL-Miloco-7B}, a 7B-parameter vision-language model tailored for home scenarios and deployed as the perception core of the Xiao\textbf{Mi} \textbf{lo}cal \textbf{co}pilot framework. Built on top of MiMo-VL-7B \cite{xiaomi2025mimo}, MiMo-VL-Miloco-7B inherits a strong general-purpose multimodal backbone and further specializes it for home environments, covering recognizing common daily activities (Play Esports, Workout, Watch TV, Reading, Play on the Phone) and understanding hand gestures (Thumbs Up, V Sign, Open Palm, OK, Shaka Sign). By connecting to Xiaomi MiJIA cameras as visual sensors and integrating with the Xiaomi IoT ecosystem, the model enables rich automation, ranging from instructions such as ``turn on the reading lamp when someone is reading'' to more complex instructions such as “if the baby cries after 10 p.m., dim the nursery lights to 10 \%, play white-noise audio on the ceiling speaker, and push a quiet notification to the parents’ phones.” \cite{ur2016trigger, perera2013context}.

The development of MiMo-VL-Miloco-7B follows a two-stage training pipeline designed to inject valuable home-scenario knowledge while preserving the base model’s generality \cite{liu2023visual, dai2023instructblip}. In the first stage, we perform supervised fine-tuning (SFT) on curated data from home environments. This stage emphasizes chain-of-thought supervision so that the model not only predicts labels but also learns explicit reasoning patterns about activities in home scenarios \cite{wei2022chain}. In addition, we adopt token-budget-aware reasoning data that encourage concise, task-oriented responses. Together, chain-of-thought supervision and token-budget-aware reasoning enable the model to learn knowledge in a data-efficient manner while also performing reasoning efficiently, which is important for latency-sensitive edge deployment \cite{zeng2024token, zheng2025review}.

In the second stage, we apply GRPO-based reinforcement learning on top of the SFT model \cite{shao2024deepseekmath, guo2025deepseek}. Leveraging the Time-R1 \cite{wang2025timer1} data construction strategy, we assemble efficient training sets across multiple domains and optimize the model with mixed reinforcement signals \cite{wang2025timer1, su2025crossing} to further enhance its perception and reasoning capabilities while avoiding over-specialization. This stage focuses primarily on improving performance on generic multimodal benchmarks \cite{yue2024mmmu, fang2024mmbench}, while maintaining strong home-scenario understanding, including gesture recognition and activity classification. Empirically, MiMo-VL-Miloco-7B achieves strong F1 scores on internal home-scenario evaluations, while also showing gains on several public multimodal understanding and reasoning benchmarks\cite{wang2025timer1, yue2024mmmu, fang2024mmbench}, with only modest degradation on tasks that are less relevant to its target use cases, such as OCR or mathematical reasoning.

Our contributions can be summarized as follows:

\begin{enumerate}
    \item \textbf{Home-centric multimodal modeling on the edge.} We introduce MiMo-VL-Miloco-7B, an edge-deployable vision-language model specifically optimized for home environments, capable of robust activity recognition, gesture understanding, and natural-language interaction with home scenarios, achieving competitive performance across multimodal benchmarks.

    \item \textbf{A two-stage tuning framework for home scenarios.}
    We design a effective two-stage training pipeline, in which the first stage performs chain-of-thought enhanced supervised fine-tuning and the second stage applies GRPO-based reinforcement learning with difficulty-aware data filtering. This training strategy substantially improves home-scenario understanding, while preserving the generalization ability of the pretrained backbone.

    \item \textbf{Open ecosystem for smart-home research and deployment.} We release both MiMo-VL-Miloco-7B and its quantized variant MiMo-VL-Miloco-7B-GGUF, along with a Gradio demo and integration into the open-source Xiaomi Miloco framework. These resources provide a practical foundation for researchers and developers to explore privacy-preserving, on-device multimodal intelligence in real-world smart-home settings.
\end{enumerate}

Taken together, these efforts advance smart-home systems beyond manual rule configuration toward genuinely context-aware, privacy-preserving multimodal agents that can continuously ``see and think'' alongside users in their daily lives. By serving as a unified, edge-deployable perception and reasoning backbone for home environments, MiMo-VL-Miloco-7B provides a practical foundation for research and large-scale deployment of vision-language copilots in real-world smart-home ecosystems.

\section{Approach}

\subsection{Supervised Fine-Tuning (SFT)}

In this section, we first present the architectural design of MiMo-VL-Miloco-7B, followed by a detailed description of the data composition and the training implementation for the supervised fine-tuning stage.

\subsubsection{Architecture of MiMo-VL-Miloco}
\begin{figure}[h!]
    \centering
    \includegraphics[width=\textwidth]{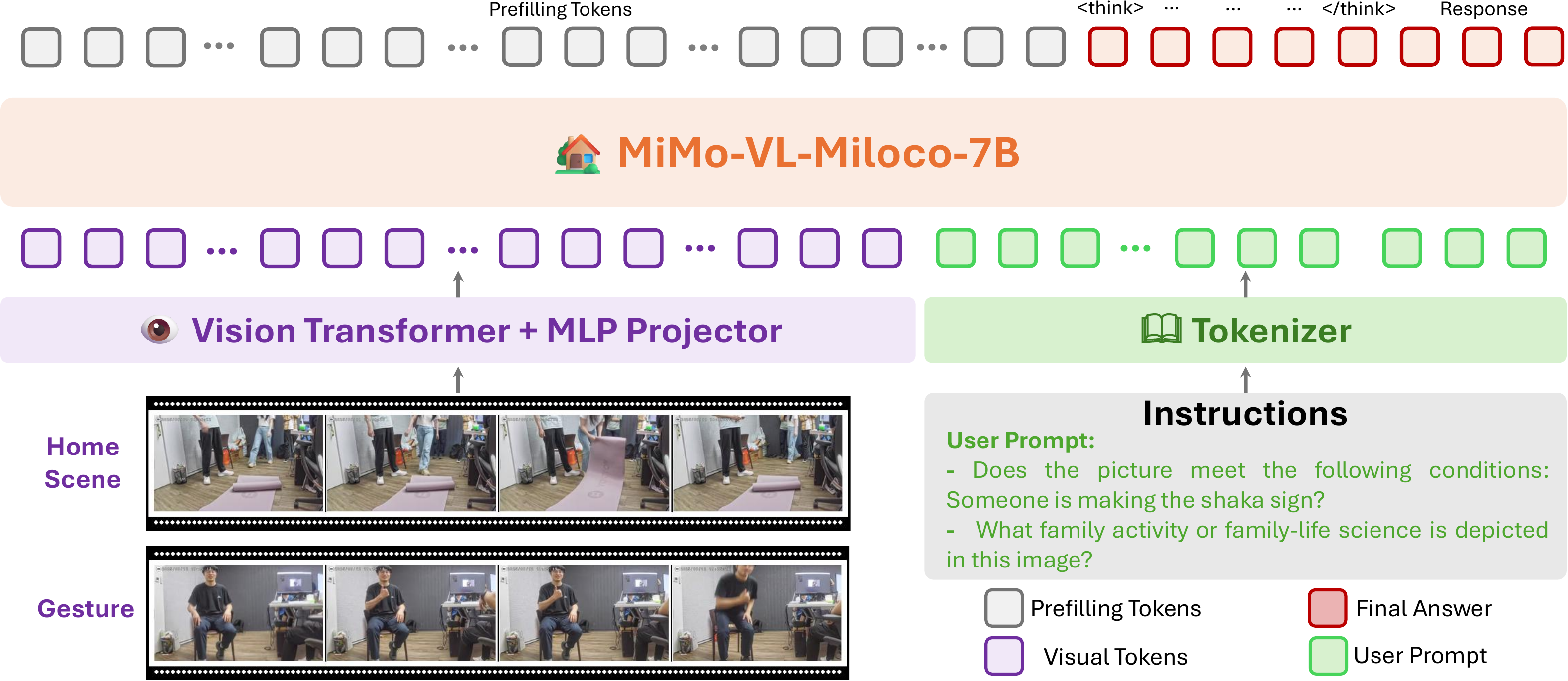}
    \caption{The overview of MiMo-VL-Miloco-7B. The model takes as input a video depicting either a home scene or user gestures. Video frames are encoded by a Vision Transformer (ViT) and projected into the LLM embedding space via an MLP projector, forming visual tokens. In parallel, the instruction prompt is tokenized into text tokens. The visual and text tokens are then concatenated and fed into the LLM backbone, which generates the final response. }
    \label{fig:arch}
\end{figure}

The architecture of MiMo-VL-Miloco-7B consists of three primary components: (1) a Vision Transformer (ViT) supporting native resolution, which encodes visual inputs; (2) a projector composed of a Multi-Layer Perceptron (MLP), serving to align visual encodings with the latent space of the Large Language Model (LLM); and (3) the LLM backbone, responsible for text comprehension and reasoning. All components, including the vision encoder, the projector, and the LLM backbone, are initialized using their respective pre-trained weights from MiMo-VL\cite{xiaomi2025mimo}, thereby inheriting its exceptional capabilities in multimodal understanding and reasoning. We show the architecture of MiMo-VL-Miloco-7B in Fig.~\ref{fig:arch}.

\subsubsection{Data}
We construct a comprehensive training set designed to enhance home-scenario understanding while maintaining general-purpose competence. Our training data comprises a carefully curated mix of proprietary home-scenario data and general multimodal data. Each subset undergoes rigorous processing to ensure high quality.

\paragraph{Home-Scenario Data.}
Addressing the scarcity of open-source data for common home scenarios and gesture recognition, we establish a strict internal data collection pipeline. Specifically, we collect data covering common daily activities including Play Esports, Workout, Watch TV, Reading, Play on the Phone and gesture recognition including Thumbs Up, V Sign, Open Palm, OK, Shaka Sign.

To ensure both data diversity and annotation quality, we define precise collection protocols for each category. Volunteers are recruited to perform the target activities and gestures under varied environmental conditions while adhering to the predefined rules. Prior to data acquisition, all participants receive standardized training to ensure a clear and consistent understanding of the protocols. During the annotation process, deviations from the protocols are corrected in real time. Finally, all annotations undergo a comprehensive review by experienced AI researchers to eliminate residual errors, resulting in a high-quality, carefully verified home-scenario dataset.

To fully exploit the informational richness of each video, we move beyond simple label prediction by generating \textbf{chain-of-thought (CoT)} annotations. To facilitate effective knowledge transfer across different scenarios, we standardize the reasoning patterns used in these CoT sequences. This knowledge-intensive reasoning data enables the model to develop a deeper understanding of home scenarios, even with limited labeled data. 

However, for edge deployment, the increased inference latency associated with long CoT generation is often prohibitive. To address this, we introduce \textbf{token-budget-aware} reasoning data for model training, which prompts the model to output direct answers without explicit reasoning steps. By integrating both chain-of-thought data and token-budget-aware data, the model learns to internalize knowledge in a data-efficient manner while maintaining efficient inference speed.

\paragraph{General Data.}
To preserve the model's general visual understanding and reasoning capabilities, we utilize the training corpus from the MiMo-VL. This corpus encompasses a wide range of data types to enhance capabilities including general images understanding, image grounding, OCR understanding, video understanding, and multimodal reasoning, organized into the following tasks:

\begin{itemize}[leftmargin=10pt]
    \item \textbf{VQA.} This strengthens the model's understanding of a broad spectrum of multimodal tasks, supporting its role as a general-purpose assistant.
    \item \textbf{Grounding.} These datasets enhance cross-modal localization across various domains such as general image grounding and GUI grounding, unlocking the model's potential as a GUI agent.
    \item \textbf{OCR.} Covering long documents, information graphics, and charts, the data significantly improves information extraction capabilities across multiple domains.
    \item \textbf{Video.} This includes dense captioning with timestamped event annotations, temporal grounding to link events to timelines, and synthetic QA data for complex reasoning. These tasks bolster the model's robust understanding of temporal dynamics in long videos.
    \item \textbf{Reasoning.} Containing rich reasoning processes, the reasoning data has been proven in MiMo-VL to significantly enhance perception and reasoning. In our experiments, incorporating an appropriate proportion of this data not only maintained general understanding but also improved performance in home-scenario understanding.
\end{itemize}

\subsubsection{Training Process}
We utilize the AdamW\cite{kingma2014adam} optimizer with a total batch size of 128 and a learning rate of $1 \times 10^{-5}$. The maximum sequence length was set to 32,768, and we applied a warmup ratio of 0.03. We performed full-parameter fine-tuning, updating all model parameters during training. By training on our combined proprietary home-scenario data and multimodal corpus from MiMo-VL, the model achieved a significant improvement in gesture and daily activity recognition, while preserving its robust general multimodal reasoning capabilities.








\subsection{Reinforcement Learning (RL)}

While the SFT stage significantly enhanced the model's proficiency in home-scenario understanding, our analysis reveals a performance degradation in general multimodal tasks, specifically video understanding, GUI grounding, and complex multimodal reasoning. This phenomenon, consistent with findings in continual learning literature\cite{cao2024generative}, suggests a trade-off where acquiring new domain-specific knowledge leads to catastrophic forgetting of previously learned capabilities. To mitigate this, we employ RL to restore general competence while ensuring that the home-scenario performance remains competitive with the SFT baseline.

\subsubsection{Data Construction}

We conduct a systematic analysis of the post-SFT model to identify domains suffering from regression and construct a targeted RL dataset accordingly. The data composition is as follows:

\begin{itemize}[leftmargin=10pt]
    \item \textbf{Video Understanding.} To ameliorate the model's grasp of temporal dynamics, we focus on the temporal grounding task. By training the model to precisely locate time segments corresponding to textual queries, we aim to improve overall video comprehension. Our data sources include VTG-IT\cite{guo2025vtg}, TimeIT\cite{ren2024timechat}, TimePro\cite{zeng2410timesuite}, HTStep\cite{afouras2023ht}, and LongVid\cite{li2024videochat}.
    \item \textbf{GUI Grounding.} We curate data from multiple sources. These samples are designed to recover and enhance the model's ability to interpret and localize elements within graphical user interfaces.
    \item \textbf{Multimodal Reasoning.} To bolster general reasoning capabilities, we select multimodal STEM problems, which include a substantial number of mathematics-related queries. 
\end{itemize}

Inspired by the Time-R1 data construction strategy~\cite{wang2025timer1}, we apply a difficulty filter to exclude samples that were either trivial or excessively difficult. This filtering strategy yields two key benefits: (1) it stabilizes training by providing consistent optimization gradients, and (2) it accelerates convergence while minimizing the interference with home-scenario knowledge retention.

\subsubsection{Training Process}

In the RL stage, we employ Group Relative Policy Optimization (GRPO), an algorithm that eliminates the need for a separate critic model by leveraging group-based advantage estimation.

\paragraph{Group Relative Policy Optimization (GRPO).}
For each query $q$, GRPO samples a group of outputs $\{o_1, o_2, \dots, o_G\}$ from the current policy $\pi_\theta$. The optimization objective is formulated as:

\begin{equation}
\begin{split}
    \mathcal{J}_{\text{GRPO}}(\theta) = \mathbb{E}_{q \sim P(Q), \{o_i\}_{i=1}^G \sim \pi_{\theta_{\text{old}}}} \Bigg[ \frac{1}{G} \sum_{i=1}^G \Bigg( \min \bigg( \frac{\pi_\theta(o_i|q)}{\pi_{\theta_{\text{old}}}(o_i|q)} \hat{A}_i, \\
    \text{clip}\left(\frac{\pi_\theta(o_i|q)}{\pi_{\theta_{\text{old}}}(o_i|q)}, 1-\epsilon, 1+\epsilon\right) \hat{A}_i \bigg) - \beta \mathbb{D}_{\text{KL}}(\pi_\theta || \pi_{\text{ref}}) \Bigg) \Bigg]
\end{split}
\end{equation}

where $\epsilon$ and $\beta$ are hyperparameters controlling the clipping range and the KL-divergence penalty, respectively. The advantage $\hat{A}_i$ for each output is computed by normalizing the rewards within the group, thereby using the group average as a baseline:

\begin{equation}
    \hat{A}_i = \frac{r_i - \text{mean}(\{r_1, \dots, r_G\})}{\text{std}(\{r_1, \dots, r_G\})}
\end{equation}

where $r_i$ is the reward for output $o_i$. This approach reduces memory overhead and training complexity compared to standard PPO with a value network.

\paragraph{Reward Modeling.}
To guide the model across diverse tasks, we design specific reward functions. For all tasks, the total reward $r$ is a weighted sum of an accuracy-based reward $r_{\text{acc}}$ and a format-compliance reward $r_{\text{fmt}}$, defined as:
\begin{equation}
    r = \lambda_{\text{acc}} \cdot r_{\text{acc}}(y, y^*) + \lambda_{\text{fmt}} \cdot r_{\text{fmt}}(y)
\end{equation}
where $y$ is the predicted output, $y^*$ is the ground truth, and we set $\lambda_{acc}=0.9$ and $\lambda_{fmt}=0.1$. The format reward $r_{\text{fmt}}(y)$ is a binary value indicating whether the output adheres to the required structure (e.g., enclosing the answer in specific tokens). The task-specific accuracy rewards are defined as follows:

\noindent\textbf{1. Temporal Grounding Reward.}
For video temporal grounding, we evaluate the overlap between the predicted time segment $T_{\text{pred}} = [t_{\text{start}}^{\text{pred}}, t_{\text{end}}^{\text{pred}}]$ and the ground truth segment $T_{\text{gt}} = [t_{\text{start}}^{\text{gt}}, t_{\text{end}}^{\text{gt}}]$. The accuracy reward is calculated using the Intersection over Union (IoU) of the 1D time intervals:
\begin{equation}
r_{\text{acc}}^{\text{temp}} = \frac{\max(0, \min(t_{\text{end}}^{\text{pred}}, t_{\text{end}}^{\text{gt}}) - \max(t_{\text{start}}^{\text{pred}}, t_{\text{start}}^{\text{gt}}))}{(t_{\text{end}}^{\text{pred}} - t_{\text{start}}^{\text{pred}}) + (t_{\text{end}}^{\text{gt}} - t_{\text{start}}^{\text{gt}}) - \max(0, \min(t_{\text{end}}^{\text{pred}}, t_{\text{end}}^{\text{gt}}) - \max(t_{\text{start}}^{\text{pred}}, t_{\text{start}}^{\text{gt}}))}.
\end{equation}

\noindent\textbf{2. GUI Grounding Reward.}
For GUI tasks, the model predicts a bounding box $B_{\text{pred}}$. The reward is determined by the 2D IoU between the predicted box and the ground truth box $B_{\text{gt}}$:
\begin{equation}
    r_{\text{acc}}^{\text{gui}} = \frac{\text{Area}(B_{\text{pred}} \cap B_{\text{gt}})}{\text{Area}(B_{\text{pred}} \cup B_{\text{gt}})}
\end{equation}
\noindent where $\text{Area}(A \cap B), \text{Area}(A \cup B)$ are the intersection and union between boxes $A$ and $B$, respectively.

\noindent\textbf{3. Mathematical Reasoning Reward.}
For mathematical and reasoning tasks, we employ an exact-match criterion based on the final answer extracted from the model's chain of thought. The accuracy reward is a binary indicator function:
\begin{equation}
    r_{\text{acc}}^{\text{math}} = \mathbb{I}[y = y^*]
\end{equation}
where $\mathbb{I}[\cdot]$ is 1 if the extracted answer matches the ground truth and 0 otherwise.

By post-training through GRPO, MiMo-VL-Miloco regains its general capabilities in video, GUI, and reasoning tasks while retaining the specialized knowledge acquired during the SFT stage.

\section{Experiments}

We evaluate the model on both home-scenario benchmarks and general-purpose benchmarks. The home-scenario benchmarks assess the model’s ability to recognize common daily activities and fine-grained hand gestures, directly reflecting its applicability to smart-home scenarios. In addition, we evaluate the model on 20 general-purpose benchmarks spanning a broad range of capabilities, including visual perception, video understanding, GUI grounding, text comprehension, reasoning, and OCR. These evaluations provide a comprehensive assessment of MiMo-VL-Miloco-7B, capturing both its domain-specific strengths and its generalization performance.

\subsection{Evaluation Settings}

For image comprehension tasks, we adopt a dynamic resolution strategy, representing each input image with up to 4,096 patches of $28 \times 28$ pixels and a maximum generation budget of 32,768 tokens. To ensure reproducibility, greedy search is used for standard visual question answering tasks.
For video benchmarks, frames are sampled at 2 FPS, with a maximum of 256 frames per input and a total context limit of 16,384 tokens. For open-ended text generation and reasoning tasks, we set the maximum number of new tokens to 32,768 and use a sampling strategy with temperature 0.6 and top-p 0.95.
We extend the LMMs-Eval framework~\cite{zhang2025lmms} to better support models capable of long-chain-of-thought (CoT) reasoning and refine task-specific evaluation logic to ensure metric consistency across all benchmarks.

\begin{table}[htbp]
\centering
\small
\setlength{\tabcolsep}{4pt}
\renewcommand{\arraystretch}{1.15}
\begin{tabular}
{>{\raggedright\arraybackslash}p{2.45cm}
 >{\centering\arraybackslash}p{1.4cm} |
 >{\centering\arraybackslash}p{1.75cm}
 >{\centering\arraybackslash}p{1.4cm}
 >{\centering\arraybackslash}p{2cm} |
 >{\centering\arraybackslash}p{1.5cm}}
\toprule
\textbf{Benchmark} & \textbf{Metrics} & \textbf{Qwen2.5-VL 7B} & \textbf{Gemini-2.5-Pro} & \textbf{MiMo-VL 7B-SFT-2508} & \textbf{MiMo-VL-Miloco-7B} \\
\midrule
\grp \multicolumn{6}{l}{\textbf{Daily Activity}} \\
Watch TV            & F1-score  & 81.2   & 93.8  & 86.0   & \textbf{98.3} \\
Reading             & F1-score  & 78.7   & 88.9  & 81.7   & \textbf{90.8} \\
Play on the phone   & F1-score  & 88.4   & 81.9  & 75.3   & \textbf{90.5} \\
Play Esports        & F1-score  & 95.2   & 95.9  & 95.2   & \textbf{99.2} \\
Workout             & F1-score  & 88.9   & 91.9  & 93.7   & \textbf{96.7} \\
\midrule
\grp \multicolumn{6}{l}{\textbf{Gesture}} \\
OK                        & F1-score  & 73.8   & 81.8  & 75.6  & \textbf{86.3}\\
Thumbs Up                 & F1-score  & 78.6   & 86.9  & 77.1  & \textbf{90.5}\\
V Sign                    & F1-score  & 67.0   & 86.7  & 86.5  & \textbf{90.3}\\
Shaka Sign                & F1-score  & 33.3   & 70.1  & 67.4  & \textbf{88.3}\\
Open Palm                 & F1-score  & 53.4   & 87.4  & 79.8  & \textbf{87.5}\\
\bottomrule
\end{tabular}

\vspace{0.3cm}
\caption{Per-category F1-score comparison for home-scenario activity and hand gesture recognition. We benchmark MiMo-VL-Miloco-7B against leading baselines, including Qwen2.5-VL-7B~\cite{qwen25vl2025techreport}, Gemini-2.5-Pro~\cite{gemini25_2025}, and MiMo-VL-7B-SFT-2508~\cite{xiaomi2025mimo}. MiMo-VL-Miloco-7B consistently achieves the highest F1-scores across all categories, demonstrating superior capabilities in understanding daily activities and fine-grained gestures within home environments.}
\label{tab:homebench}
    \vspace{-20pt}
\end{table}

\subsection{Home-Scenario Understanding Capabilities}

Table~\ref{tab:homebench} reports per-category F1-scores for home-scenario activity classification and gesture recognition. Overall, MiMo-VL-Miloco-7B achieves state-of-the-art performance across all categories, substantially outperforming both open-source baselines such as Qwen2.5-VL-7B~\cite{qwen25vl2025techreport}, MiMo-VL-7B-SFT-2508~\cite{xiaomi2025mimo} and the proprietary Gemini-2.5-Pro~\cite{gemini25_2025}.

For common daily activities, MiMo-VL-Miloco-7B attains F1-scores of 98.3, 90.8, 90.5, 99.2, and 96.7 on Watch TV, Reading, Play on the Phone, Play Esports, and Workout, respectively, improving upon the strongest baselines by 2-4 percentage points in each case.

In gesture recognition, the model demonstrates high precision, achieving F1-scores of 86.3 ({OK}), 90.5 ({Thumbs Up}), 90.3 ({V Sign}), 88.3 ({Shaka Sign}), and 87.5 ({Open Palm}). The improvements are particularly notable for complex gestures. For example, on the {Shaka Sign}, MiMo-VL-Miloco-7B surpasses the best baseline by over +18 F1 points. These results confirm that our home-centric training strategy, combining high-quality proprietary data with chain-of-thought supervision, enables superior perception of daily activities and fine-grained gestures in realistic home environments.

\begin{table}[htbp]
\centering
\small
\setlength{\tabcolsep}{4pt}
\renewcommand{\arraystretch}{1.15}
\begin{tabular}
{>{\raggedright\arraybackslash}p{2.45cm}
 >{\centering\arraybackslash}p{1.4cm} |
 >{\centering\arraybackslash}p{1.4cm}
 >{\centering\arraybackslash}p{1.75cm}
 >{\centering\arraybackslash}p{1.4cm}
 >{\centering\arraybackslash}p{2cm} |
 >{\centering\arraybackslash}p{1.5cm}}
\toprule
\textbf{Benchmark} & \textbf{Metrics} & \textbf{Gemma-3 27B-IT} & \textbf{Qwen2.5-VL 7B} & \textbf{InternVL3 8B} & \textbf{MiMo-VL 7B-SFT-2508} & \textbf{MiMo-VL-Miloco-7B} \\
\midrule
\grp \multicolumn{7}{l}{\textbf{General}} \\
MMMU-Pro$_{\text{standard}}$     & Acc.  & 37.8    & 34.7   & 45.6  & 43.0   & \textbf{55.7}\\
MMMU-Pro$_{\text{vision}}$       & Acc.  & 24.9    & 29.4   & 37.8  & 36.4   & \textbf{47.2}\\
MMMU$_{\text{val}}^{\dagger}$    & Acc.  & 64.9    & 58.6   & 62.7  & \textbf{68.7}   & 66.4\\
MME                              & Score & -       & -      & -     & \textbf{2473.6}$^\ddagger$ & 2396.4$^\ddagger$\\
MME-RealWorld$_{\text{en}}$      & Acc.  & 51.9    & 57.4   & 56.1  & 59.7$^*$   & \textbf{59.9}$^*$ \\
MME-RealWorld$_{\text{cn}}$      & Acc.  & 47.9    & 51.2   & 58.5  & \textbf{64.3}$^*$   & 61.4$^*$ \\
\midrule
\grp \multicolumn{7}{l}{\textbf{Video}} \\
Video-MME                        & Acc.  & -       & 65.1   & 66.3  & 67.8  & \textbf{68.0}\\
Video-MMMU                       & Acc.  & -       & 47.4   & 48.9  & 57.6  & \textbf{63.6}\\
Charades-STA                     & Acc.  & -       & 43.6   & 25.4  & 44.3  & \textbf{46.6}\\
\midrule
\grp \multicolumn{7}{l}{\textbf{Doc \& OCR}} \\
ChartQA$^\dagger$                & Acc.  & 78.0    & 90.2   & 89.6  & \textbf{93.2}  & 92.0\\
DocVQA$^\dagger$                 & Acc.  & 86.6    & 95.5   & 89.4  & \textbf{95.5}  & 95.2\\
OCRBench$^\dagger$               & Acc.  & 77.6    & \textbf{89.7}  & 88.0  & 87.8   & 86.5\\
\midrule
\grp \multicolumn{7}{l}{\textbf{GUI}} \\
ScreenSpot                       & CenterAcc. & -       & 84.7      & 79.5  & 89.5  & \textbf{89.8}\\
ScreenSpot-v2                    & CenterAcc. & -       & 88.0      & 81.4  & 92.0  & \textbf{92.1}\\
ScreenSpot-Pro                   & CenterAcc. & -       & 29.0      & -     & 37.3  & \textbf{37.8}\\
\midrule
\grp \multicolumn{7}{l}{\textbf{Reasoning}} \\
OlympiadBench                    & Acc.       & -       & -       & -       & \textbf{57.9}  & \textbf{57.9}\\
MathVision                       & Acc.       & -       & -       & -       & \textbf{57.2}  & 54.0\\
\midrule
\grp \multicolumn{7}{l}{\textbf{Text}} \\
MMLU-Pro                         & EM    & 45.3   & 48.7   & 58.3  & 67.1  & \textbf{68.5}\\
GPQADiamond                      & Pass@1& 40.9   & 30.3   & 33.5  & \textbf{55.9}  & 55.1\\
MATH500                          & Pass@1& 29.3   & 25.4   & 29.4  & \textbf{96.8}  & 95.2\\
\bottomrule
\end{tabular}

\vspace{0.3cm}
\caption{Performance comparison on general-purpose multimodal benchmarks. We evaluate MiMo-VL-Miloco-7B against leading open-source baselines, including Gemma-3-27B-IT~\cite{team2025gemma}, Qwen2.5-VL-7B~\cite{qwen25vl2025techreport}, InternVL3-8B~\cite{zhu2025internvl3}, and the base model MiMo-VL-7B-SFT-2508~\cite{xiaomi2025mimo}. The assessment encompasses a diverse range of domains: general vision-language tasks, video understanding, document/OCR, GUI grounding, and text-based reasoning. $\dagger$: Evaluated using GPT-4o as the judge. $*$: Results obtained without explicit chain-of-thought reasoning. $\ddagger$: Results reproduced using our evaluation pipeline. The best performance among open-source models is highlighted in \textbf{bold}.}
\label{tab:benchmarks}
    \vspace{-20pt}
\end{table}

\subsection{General Capabilities}

Table~\ref{tab:benchmarks} presents the evaluation of MiMo-VL-Miloco-7B on general multimodal benchmarks, compared with leading open-source vision-language models. Overall, the model demonstrates strong performance across image, video, GUI, and reasoning tasks. While the base model, MiMo-VL-7B-SFT-2508, maintains a slight edge in specialized document understanding and OCR tasks, MiMo-VL-Miloco-7B effectively balances home-scenario specialization with general competence.
The specific findings are as follows:

\noindent\textbf{Image Understanding.} MiMo-VL-Miloco-7B achieves 55.7\% and 47.2\% accuracy on MMMU-Pro\textsubscript{standard} and MMMU-Pro\textsubscript{vision}~\cite{yue2025mmmu}, respectively, outperforming larger models such as Gemma-3-27B-IT~\cite{team2025gemma} and InternVL3-8B~\cite{zhu2025internvl3} by significant margins. It also secures the top score on MME-RealWorld\textsubscript{en} (59.9\%) while remaining highly competitive on MME-RealWorld\textsubscript{cn}~\cite{zhang2024mme}.

\noindent\textbf{Video Understanding.} Reflecting the benefits of our RL stage, the model sets new state-of-the-art scores among compared models with 63.6\% on Video-MMMU~\cite{hu2025video} and 46.6\% on Charades-STA~\cite{Gao_2017_ICCV}. Notably, it shows improvement on Video-MME~\cite{fu2025video} compared to the SFT baseline.

\noindent\textbf{Document and OCR.} MiMo-VL-Miloco-7B leads the open-source comparisons with 92.0\% on ChartQA~\cite{masry2022chartqa} and 95.2\% on DocVQA~\cite{mathew2021docvqa}. Although it trails slightly behind the base MiMo-VL-7B-SFT-2508, it significantly outperforms other open-source large models, indicating that the trade-off for home specialization is minimal.

\noindent\textbf{GUI Grounding.} The model achieves new record highs in GUI agents tasks, recording 89.8\%, 92.1\%, and 37.8\% CenterAcc on ScreenSpot~\cite{cheng2024seeclick}, ScreenSpot-v2~\cite{cheng2024seeclick}, and ScreenSpot-Pro~\cite{cheng2024seeclick}, respectively.

\noindent\textbf{Text and Reasoning.} In text-only reasoning, MiMo-VL-Miloco-7B achieves an Exact Match (EM) score of 68.5 on MMLU-Pro~\cite{wang2024mmlu}, significantly surpassing other open-source models. Furthermore, it attains 95.2 Pass@1 on MATH500~\cite{hendrycks2021measuring} and 55.1 Pass@1 on GPQA-Diamond~\cite{rein2024gpqa}, performing on par with the best results from the MiMo-VL family. These results demonstrate that our home-centric specialization preserves, and in some cases enhances, general language and reasoning capabilities through the GRPO-based refinement.

In summary, compared with mainstream open-source models, MiMo-VL-Miloco-7B demonstrates clear advantages in video understanding, GUI grounding, and multimodal reasoning while maintaining competitive performance in text reasoning tasks. The minor regression in document-centric tasks is expected due to home-scenario prioritization, but the model remains robust for practical deployment. Future work will focus on bridging this gap through multi-objective optimization.

\section{Conclusions}


In this work, we present \textbf{MiMo-VL-Miloco-7B}, demonstrating that targeted home-centric tuning can effectively coexist with broad multimodal competence. By integrating enhanced supervised fine-tuning with GRPO-based reinforcement learning utilizing the difficulty-aware data construction strategy, the model acquires rich priors regarding household activities and gestures while preserving robust performance on generic video, GUI, text, and reasoning benchmarks. Extensive evaluations confirm substantial F1 gains on internal home-scenario suites and competitive results across public multimodal leaderboards, with only modest trade-offs on document-centric tasks. Through the joint release of full-precision and quantized GGUF checkpoints  and integration with the framework of Xiaomi Miloco , we provide a reproducible foundation for privacy-preserving, on-device multimodal agents capable of perceiving, reasoning, and acting alongside users in real-world smart homes.

As future work, we will further capitalize on Xiaomi’s hardware ecosystem to incorporate a broader range of sensing modalities, including audio and millimeter-wave signals, into a unified multimodal learning framework. By jointly reasoning over heterogeneous sensory inputs, we aim to advance toward holistic household scene understanding and fine-grained spatial perception. Meanwhile, we will continue to investigate model compression and architectural optimization to achieve smaller model footprints and more efficient inference, facilitating scalable and practical on-device deployment in real-world smart home scenarios.
\bibliographystyle{plainnat}    
\bibliography{refs}             


\appendix

\section{Contributions and Acknowledgments}

We express our sincere gratitude to all contributors for their dedication to the design, implementation, experimentation, and documentation of this project. Their collaborative efforts across data curation, model training, system integration, and empirical evaluation were instrumental to the success of this work and the release of the accompanying codebase. We extend special thanks to the Xiaomi LLM-Core and CloudML teams for their invaluable support in dataset construction, training infrastructure, and evaluation protocols. The contributors to this work are listed as follows:

\begin{multicols}{2}
\noindent
\textbf{Core Contributors}\\
Jiaze Li\\
Jingyang Chen\\
Yuxun Qu\\
Jianzhong Ju\\
Zhenbo Luo \\
Jian Luan\\

\columnbreak

\noindent
\textbf{Contributors}\\
Shijie Xu\\
Zhenru Lin \\
Junyou Zhu \\
Boshen Xu \\
Wenhui Tan \\
Pei Fu \\
\end{multicols}



\end{document}